\documentclass[letterpaper, 10 pt, conference]{ieeeconf}  

\IEEEoverridecommandlockouts                              

\makeatletter
\let\NAT@parse\undefined
\makeatother

\usepackage[noadjust]{cite}
\usepackage{amsmath,amssymb,amsfonts}
\usepackage{graphicx}
\usepackage{dblfloatfix}
\usepackage{bm}
\usepackage{siunitx}
\usepackage{booktabs}
\usepackage{hyperref}
\usepackage{cleveref}

\Crefname{equation}{Eq.}{Eqs.}
\Crefname{figure}{Fig.}{Figs.}
\Crefname{tabular}{Tab.}{Tabs.}

\DeclareMathOperator{\arctantwo}{arctan2}

\title{\LARGE \bf
Direct learning of home vector direction\\for insect-inspired robot navigation
}

\author{Michiel V.M. Firlefyn, Jesse J. Hagenaars and Guido C.H.E. de Croon
\thanks{This publication is part of NWA ACT, which is funded by NWO (NWA.1292.19.298). All authors are with the Micro Air Vehicle Laboratory,  Faculty of Aerospace Engineering, Delft University of Technology, The Netherlands. Correspondence: {\tt j.j.hagenaars@tudelft.nl}.}%
}

\begin{document}

\maketitle
\thispagestyle{empty}
\pagestyle{empty}

\begin{abstract}
Insects have long been recognized for their ability to navigate and return home using visual cues from their nest's environment. However, the precise mechanism underlying this remarkable homing skill remains a subject of ongoing investigation. Drawing inspiration from the learning flights of honey bees and wasps, we propose a robot navigation method that directly learns the home vector direction from visual percepts during a learning flight in the vicinity of the nest. After learning, the robot will travel away from the nest, come back by means of odometry, and eliminate the resultant drift by inferring the home vector orientation from the currently experienced view. Using a compact convolutional neural network, we demonstrate successful learning in both simulated and real forest environments, as well as successful homing control of a simulated quadrotor. The average errors of the inferred home vectors in general stay well below the $90^\circ$ required for successful homing, and below $24^\circ$ if all images contain sufficient texture and illumination. Moreover, we show that the trajectory followed during the initial learning flight has a pronounced impact on the network's performance. A higher density of sample points in proximity to the nest results in a more consistent return. Code and data are available at \url{https://mavlab.tudelft.nl/learning_to_home}.
\end{abstract}

\section{INTRODUCTION}

\begin{figure}[t!]
    \centering
    \includegraphics[width=0.94\linewidth]{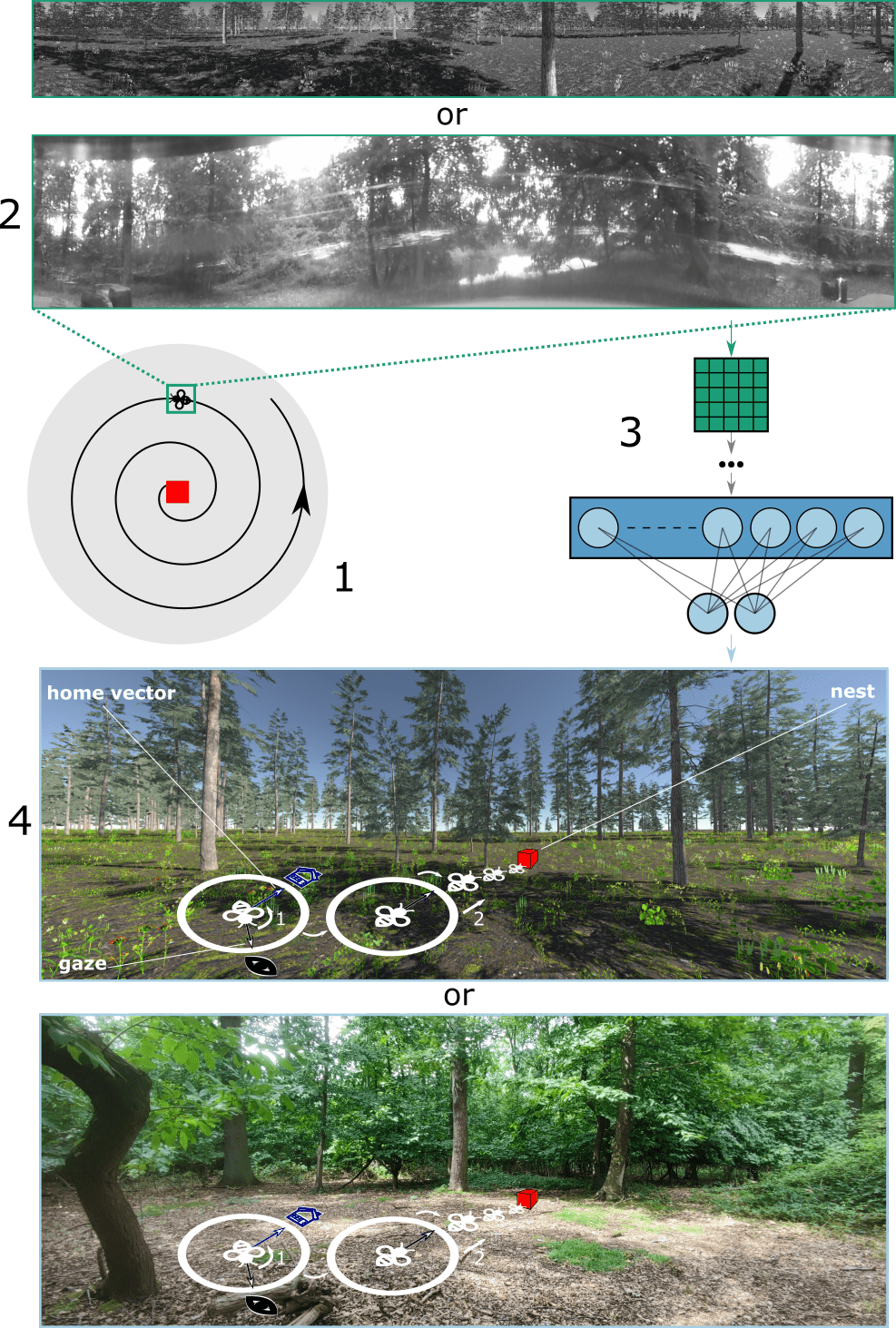}
    \caption{During a learning trajectory (1), we capture omnidirectional images (2) and train a convolutional neural network (3) to estimate the home vector direction in a simulated or real environment (4).}
    \label{fig:abstract}
\end{figure}

Imagine yourself in the middle of a forest. How accurately can you point to the direction of your home without external aids? Flying insects like honey bees and wasps have evolved into experts regarding such high-level navigation tasks,
with foraging trips covering distances of up to $13.5$~km from their nests~\cite{vonfrisch1993dance}. Throughout such trips, they retain a sense of the home direction through path integration~\cite{haferlach2007evolving} based on the Sun's polarization pattern and the optical flow across the retinal image~\cite{srinivasan1996honeybee}.
To ensure consistent returns, any odometric drift accumulated over this journey will have to be corrected for.
To do so, honey bees emerging from the nest for the first time perform `learning' flights around the hive, acquiring visual memories that can be interpreted in later flights in order to successfully navigate back to the hive~\cite{capaldi2000ontogeny,zeil2012visual}.

Two main models of insect-inspired navigation that could accomplish this have been proposed. In the `snapshot model'~\cite{cartwright1983landmark} a stored snapshot view is compared with the current view, and the robot moves to minimize their difference~\cite{lambrinos2000mobile,hafner2001learning}. The robot can determine a direction home if the views contain sufficient similar elements for matching, restricting homing success to a limited \emph{catchment area} around the snapshot. For navigating larger distances, snapshots can be linked together in a sequence, allowing for route following. The `familiarity model' is an alternative model for route following that proposes to move into the most familiar direction when returning to the nest~\cite{baddeley2012model}. Typically, a neural network is trained to discriminate between views that have been perceived before and views that are novel. Various methods, from rotating physically to using repulsion and attraction, have been proposed to find the most familiar direction on the way back to the nest ~\cite{baddeley2012model,lemoel2020opponent,gattaux2023antcar}. Route following has two main problems: (i) getting off-route, e.g., by one failure in homing to a snapshot or determining the familiar direction, can lead to a complete navigation loss, and (ii) the outbound route may be very tortuous and hence suboptimal for following as an inbound route. 

Instead of route following, many insects return to their nest in a straight trajectory on a novel path \cite{wehner1990insect,webb2019internal}. In \cite{webb2019internal} it has been argued that insects navigate by means of a nest-centered coordinate frame, always maintaining an internal representation of a vector to the nest. This view is also supported by the fact that honey bees convey this type of information to other bees when communicating about the location of flower fields through waggle dances~\cite{chittka2022mind}.   

Interestingly, honey bees perform `learning flights' of increasing area when they first leave the hive \cite{degen2015exploratory}. After a few learning flights, they are able to navigate to far-away flower fields and return successfully. We hypothesize that honey bees use these learning flights to learn a \emph{direct} mapping from visual percepts to a vector representation of home, consisting of direction and possibly distance. Based on this hypothesis, we introduce a novel insect-inspired navigation algorithm that uses learning flights around the home to learn how to directly map the currently experienced view to the home direction.

This algorithm can be abstractly implemented on a robot, making use of the available insect sensor suite, but not implementing their specific neural architecture. More specifically, we propose to train a compact convolutional neural network (CNN) that takes rectified catadioptric omnidirectional images around the nest as input and returns a unit vector representing the relative home direction. \Cref{fig:abstract} shows an overview of this.
After learning, a robot can travel far outside the range of its learning flights and return based on odometry. Upon return, the trained visual homing network can compensate for any odometric drift as long as this drift falls within the region explored during the learning flights.
We demonstrate successful learning in both simulated and real forest environments, as well as successful homing control of a simulated quadrotor.
This approach is very promising for navigation in large environments, since the learning flight effectively creates a very large homing catchment area. Depending on the odometry drift, robots can travel much farther than this area's diameter. Moreover, in contrast to previous approaches, our method allows robots to travel back home with odometry on straight, previously unseen paths. Finally, it is a very memory efficient navigation method as the homing knowledge is stored in a relatively small neural network.

\section{RELATED WORK}

The first known model to consider insect-inspired navigation is the snapshot model by Cartwright and Collett~\cite{cartwright1983landmark}. Based on their own experiments, which involved differing arrays of landmarks, they observed that visual homing in honey bees relies on visual memories acquired at the nest location, with the bees primarily concerned with the apparent size and relative bearing of landmarks. They theorized a framework where the currently perceived horizontal view is compared to the remembered one, resulting in a home vector that is oriented towards the hive.

Lambrinos et al.~\cite{lambrinos2000mobile} conducted a robotic experiment of the snapshot hypothesis using a wheeled robot equipped with an omnidirectional camera. Their set-up consisted of black cylinders positioned on a flat $40\times40$~m desert terrain.
Although their tested variants address many limitations of the original snapshot model, they still face challenges when visual noise or distant landmarks are introduced to the testing area~\cite{cartwright1983landmark}. Cues that have a large impact on home vector direction, such as landmark occlusion or mismatches, appear to be the most common obstacles.

Hafner and M\"oller~\cite{hafner2001learning} demonstrated the efficiency and simplicity of learning the home vector by visually guiding a wheeled robot to its target location in a simple simulated environment with black cylinder landmarks against a white background. They employed a snapshot-inspired approach, where a multi-layer perceptron (MLP) was trained to produce a population encoding, representing $8$ possible homing directions, based on concatenating the current and home views as inputs to the network. Instead of concatenating multiple views, we propose to learn from only the currently experienced view, in more complex environments.

Baddeley et al.~\cite{baddeley2012model} developed a route-following framework that instead learns a familiarity score based on acquired images along a route. By following the images with a relatively high familiarity score, the route can be retraced. While scene familiarity can account for changing landscapes and distant landmarks, it alone cannot provide an accurate indication of the nest direction without a continual stop-scan-go routine.
This could be addressed by incorporating a repulsive score, which, when combined with the attractive score, yields an angular error indicating the most familiar viewing direction~\cite{lemoel2020opponent}. Van Dalen et al.~\cite{vandalen2018visual} presented a validation of scene familiarity using data obtained from a simulated drone, while Gattaux et al.~\cite{gattaux2023antcar} implement familiarity-based control on a small car-like robot. Although the concept of attractive/repulsive scene familiarity appears to resolve previous limitations of homing algorithms, one could question the necessity of inferring two abstract measures from an image and subsequently integrating them into a relative angle that indicates the local home vector bearing. Alternatively, the same outcome could be achieved by directly learning the bearing from the input image. 

Consequently, we present the first visual homing algorithm that allows a robot to learn the home directions directly during its learning flight, while only requiring the currently experienced omnidirectional view and odometry to return to its starting location.

\section{METHODOLOGY}

As shown in \Cref{fig:abstract}, we perform experiments in both simulated and real forest environments, collecting omnidirectional images in regular-grid or spiral patterns around a nest location. Using these images, we train a compact CNN in a supervised manner to estimate the home vector direction relative to the current gaze. In simulation, we investigate the generalization to unseen locations, and use the trained network for homing of a simulated quadrotor. The following subsections explain these parts in more detail.

\subsection{Environment}

We use Flightmare~\cite{song2021flightmare} and Unity to create a $1000\times1000$~m photo-realistic forest environment with grass, flowers, different kinds of bushes and trees, and Perlin-noise terrain elevation (\Cref{fig:scene}). We spawn quadrotors in the desired pattern at a fixed height of $1.89$~m above the terrain, and capture six images (front, back, left, right, dorsal, ventral) with a $90^\circ$-FoV camera with a resolution of $1024\times1024$ pixels. These are then combined into a single image using a catadioptric projection~\cite{berenguel-baeta2020omniscv} with a latus rectum and distance to the camera image plane of $0.1$~m. To get an omnidirectional view, we rectify the image from polar to Cartesian coordinates, setting the gaze angle for this rectification as the angle where the circular image is cut.

The position of the nest is set at $[538,573]^T$. Images are captured in either a rectangular grid or an Archimedean spiral (representing the honey bee's learning flights). Depending on the experiment, the $10\times10$~m grid is either spaced at $1$~m or $0.25$~m intervals. The spiral follows radius $r = 0.25\theta$, with $\theta\in[0,8\pi]$ in steps of $0.08\pi$.

We assume two-dimensional movement at the fixed height of $1.89$~m to simplify rotations and avoid introducing tilt in the acquired images. Furthermore, for labeling and control purposes, we assume a source of heading information is available, such as a magnetic or polarization compass, as has been observed in bees~\cite{gould1978bees,rossel1986polarization}. Lastly, as we focus purely on navigation, we do not consider collisions, and allow simulated quadrotors/cameras to move and spawn in objects, possibly leading to (partial) occlusions of the view.

\subsection{Network architecture and training}

\begin{figure}[t!]
    \centering
    \includegraphics[width=\linewidth]{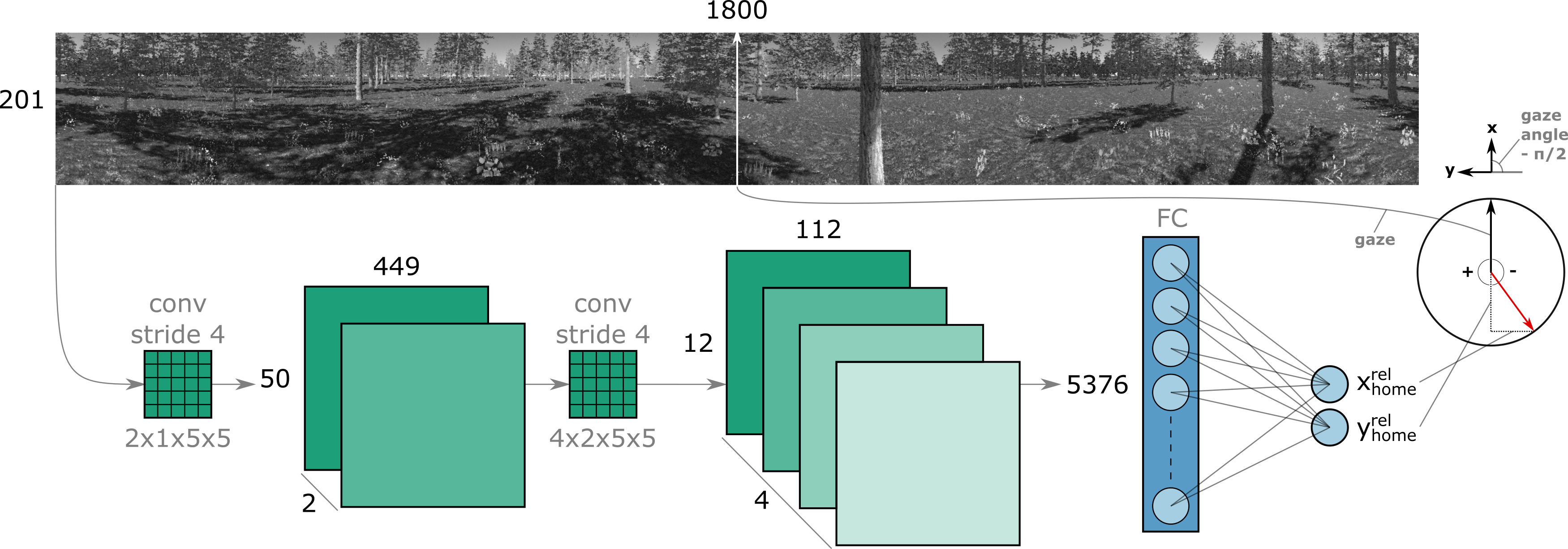}
    \caption{CNN with kernel and feature map sizes and example input. Convolutional layers are shown in teal; fully connected (FC) layers in blue. The gaze is indicated as a black or white arrow and the home vector as a red arrow. The last nodes have an output value in the range of $[-1,1]$, encoding the coordinates of the relative home vector. Note that the presented gaze angle $0^\circ$ corresponds to north/upwards.}
    \label{fig:cnn}
\end{figure}

We employ the compact three-layer CNN shown in \Cref{fig:cnn}, which consists of two convolutional layers and one fully connected layer. The input image, grayscale with a shape of $1\times201\times1800$ (channels$\times$height$\times$width), is convolved twice with $5\times5$ kernels and a stride of $4$, doubling the channel count each time. The resulting $4\times12\times112$ feature map is flattened and passed through a fully connected layer with two outputs, which represent the x- and y-coordinate of the egocentric home vector relative to the subject's gaze on the unit circle. All layers utilize the hyperbolic tangent ($\tanh$) activation function in their respective intermediate outputs. The resulting network consists of $11$k parameters (42~kB).

Directly predicting the direction to the nest as a single relative angle did not give good results, likely because similar angles might have very different values (e.g., $-175^\circ$ and $175^\circ$). To resolve this, we decompose the relative angle into its two-dimensional components on the unit circle, as can be seen in \Cref{fig:cnn}. Note that the angle formed by the home vector coordinates is relative to the subject's gaze, i.e., the middle of the currently experienced omnidirectional view.

To train the network, we rectify every omnidirectional image collected during the learning trajectory into $360$ distinct gaze directions ($1^\circ$ spacing) and compute the ground truth relative home vectors $\bm{x}^\mathrm{rel}_\mathrm{home}$ from geometric relations for all these gazes:
\begin{align}
    \bm{x}^\mathrm{rel}_\mathrm{home} &= R_z (-\omega_\mathrm{gaze})\bm{x}_\mathrm{home} \label{eq:coords} \\
    R_z(-\omega_\mathrm{gaze}) &= 
    \begin{bmatrix}
        \cos\left(\omega_{\mathrm{gaze}}\right) & \sin\left(\omega_{\mathrm{gaze}}\right) \\
        -\sin\left(\omega_{\mathrm{gaze}}\right) & \cos\left(\omega_{\mathrm{gaze}}\right)
    \end{bmatrix}
\end{align}

where $\bm{x}_\mathrm{home}$ is the home vector relative to north ($[0,1]^T$) and $\omega_\mathrm{gaze}$ is the gaze angle relative to north. The latter can be computed from the gaze vector $\bm{x}_\mathrm{gaze}$ and north vector using $\arctantwo$. For training, we use the mean squared error on the predicted and ground-truth relative home vector as loss function. Furthermore, we use a batch size of $1$ and a learning rate of \num{9e-4}, and train for a single epoch. Note that these settings were chosen as a step towards online learning. All training is done in PyTorch on CPU.\footnote{Intel Core i7-8550U running at 1.80 GHz.}

As described, training trajectories are either a $10\times10$~m grid or an Archimedean spiral. With $100$ sample locations and $360$ rectified images per location, we get a total of $35,640$ training images (excluding the nest location), which we shuffle before training. For all results in the next section, we evaluate the performance for a single northward gaze angle, as the input to the network has to be rectified, and we cannot show results for all $360$ possible gaze angles. \Cref{ssec:control} is an exception: in that case, we control the quadrotor's gaze/heading based on the network's output.

\section{EXPERIMENTS}

\subsection{Learning and generalization in simulation}
\label{ssec:generalization}

\begin{figure}[t!]
    \centering
    \includegraphics[width=\linewidth]{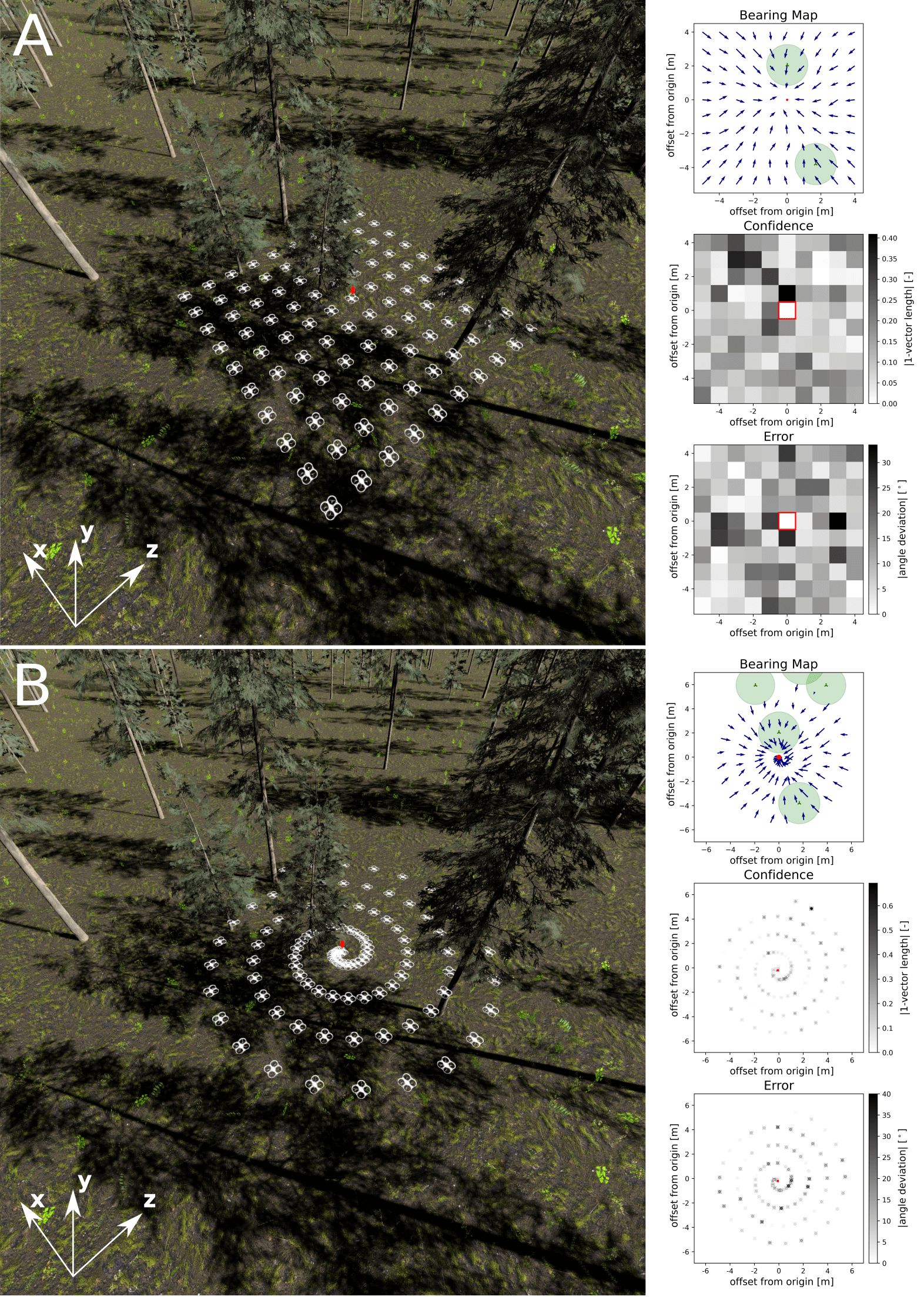}
    \caption{Simulation environment and training performance for (A) a $10\times10$~m grid and (B) an Archimedean spiral. The bearing maps show the predicted home vectors for all sampled locations, with the nest location in red and trees shaded in green.}
    \label{fig:scene}
\end{figure}

Training on the $10\times10$~m grid and Archimedean spiral flight patterns, we get the performance as shown in \Cref{fig:scene}. The bearing map shows the network's predicted relative home vector $\bm{\hat x}^\mathrm{rel}_\mathrm{home}$ at every location. If this home vector is not a unit vector, this suggests that the network has not converged to an exact solution, even if the predicted angle is correct. We interpret this as the `confidence' of the network in its prediction. The prediction error, or angle deviation, is computed as the absolute difference between the predicted and ground-truth relative home direction, $\hat\omega$ and $\omega$, which are computed from $\bm{\hat x}^\mathrm{rel}_\mathrm{home}$ and $\bm{x}^\mathrm{rel}_\mathrm{home}$ with $\arctantwo$.

Training the network on images captured in a rectangular grid yields a better overall performance compared to the spiral case, with respective average errors\footnote{We report average errors in degrees instead of the mean squared error in m\textsuperscript{2} used during training to allow better interpretation.} of $15.54^\circ$ and $19.72^\circ$. However, it is worth noting that, in the case of control, the robot should be able to return to the vicinity of the nest in both scenarios, since the angular errors of all locations are below $45^\circ$. Furthermore, looking at the shaded green areas in the bearing maps in \Cref{fig:scene}, areas near trees that are surrounded by foliage do not show a significant decline in performance. It seems that important landscape features remain interpretable, even when they are partially occluded.

\begin{table}[t!]
    \centering
    \caption{Average error $[^\circ]$ when evaluating on high-resolution grid.}
    \label{tab:generalization}
    \begin{tabular}{@{}rcc@{}}
    \toprule
                               & \textbf{Trained w. label noise} & \textbf{Trained w.o. label noise} \\ \midrule
    \textbf{Trained on grid}   & $16.82 \pm 18.98$               & $17.66 \pm 18.43$                 \\
    \textbf{Trained on spiral} & $23.34 \pm 21.77$               & $22.77 \pm 21.94$                 \\ \bottomrule
    \end{tabular}
\end{table}

\begin{figure}[t!]
    \centering
    \includegraphics[width=0.95\linewidth]{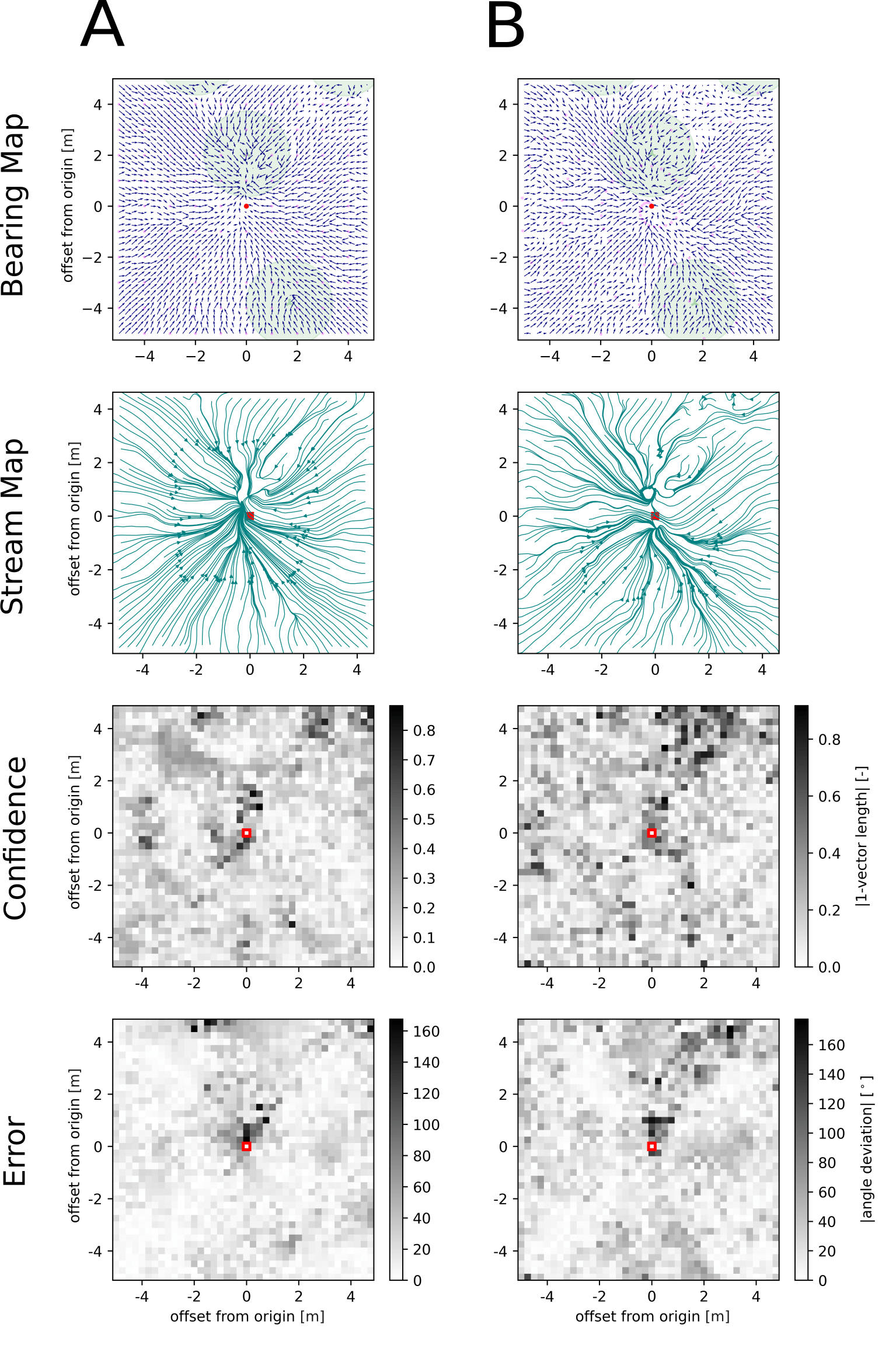}
    \caption{Generalization of networks trained on (A) a $10\times10$~m grid and (B) an Archimedean spiral to unseen locations, showing predicted home vectors and resulting stream plots. Locations seen during training are in magenta, the nest is in red and trees are shaded in green.}
    \label{fig:generalization}
\end{figure}

To demonstrate generalization to unseen locations, we capture images in a grid with decreased spacing ($0.25$~m instead of $1$~m), while maintaining the $10\times10$~m grid size. \Cref{fig:generalization} shows the performance when evaluating on this higher-resolution grid after training on the lower-resolution grid or the spiral trajectory. Once again, training on the grid exhibits superior performance, as can be seen in \Cref{tab:generalization}. When the home vectors are interpreted as velocity vectors and integrated over the evaluation grid, the stream plots shown in \Cref{fig:generalization} can be generated. These plots offer insights into the potential trajectory a controlled robot would take. As such, their endpoints are a rendition of the algorithm's catchment area, delineating the boundaries for successful homing. Note that these homing streams converge toward a point that does not precisely coincide with the nest location, as this was excluded in training. Interestingly, in the case of the spiral (\Cref{fig:generalization}B), this convergence point lies closer to the hive in all cases. This difference is a result of the higher training location density near the nest. Moreover, a circular convergence area takes shape instead of a single point. Such an area creates less of an `artificial' nest location, as can be seen in the singular convergence point of the grid case.

While occlusions due to trees do not seem to affect training performance very much (\Cref{fig:scene}), they do seem to impact generalization. Looking at the bearing map and error plots in \Cref{fig:generalization}, we see that whenever the network evaluates a particular view occluded by tree foliage (home vectors located in green shading) not encountered during training, the error will be larger. Regions with high error in the plots of \Cref{fig:generalization} can therefore be attributed to the presence of occluding trees. The same holds for the `confidence' of the network's prediction, which seems to correlate to some extent with the error and the tree locations.

To examine the robustness of the algorithm, we train another set of networks with labels perturbed by Gaussian noise with $\sigma=10^\circ$. The results in \Cref{tab:generalization} show that this has a negligible impact on the evaluation performance.

\subsection{Characterizing network activation}

Existing insect-inspired visual homing approaches explicitly define which landmarks or features to use. On the other hand, we employ a learning approach without the explicit construction of such features. To get insight into what information the network uses for prediction, we construct a simplified environment consisting of flat terrain and three trees of identical appearance (the `landmark array'), and evaluate the activations of networks trained in this environment.

\begin{figure}[t!]
    \centering
    \includegraphics[width=\linewidth]{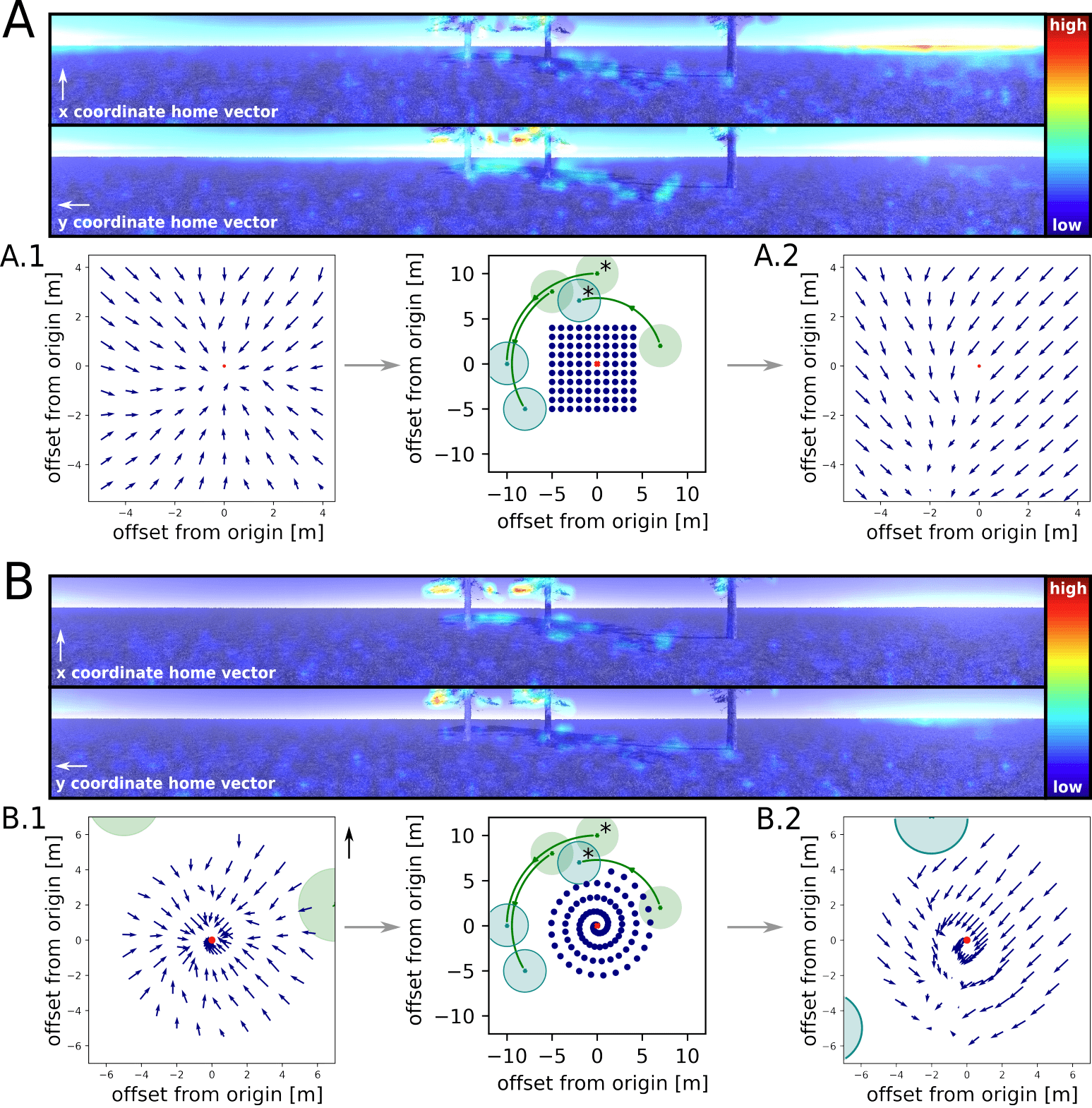}
    \caption{Analysis (following~\cite{selvaraju2017gradcam}) of the highest gradients of the network output with respect to the activations of the second convolutional layer when trained on (A) a $10\times10$~m grid and (B) an Archimedean spiral. Predicted home vectors after training (A.1/B.1) change when evaluated in an environment with a landmark array rotated $90^\circ$ CCW (A.2/B.2). Original landmarks are in green, rotated ones in teal, and dominant landmarks are marked with an asterisk.}
    \label{fig:gradcam}
\end{figure}

\Cref{fig:gradcam}A/B shows the largest-magnitude gradients of the network's output with respect to the activations of the second convolutional layer~\cite{selvaraju2017gradcam}. Areas in bright yellow and red show that the network mainly looks at a combination of horizon and foliage, as well as some tree shadows. There seems to be little difference in cues for forward/backward (x-coordinate) and left/right (y-coordinate) motion. While using trees as landmarks is not unlike honey bees, it is interesting that the networks trained here seem to consider the contrast of trees against the sky as more useful than the tree stems. This could be a result of the relatively low-resolution grayscale input, lacking diversity in shades of green compared to real-world environments, or the fact that the appearance of the sky is constant in simulation.

Furthermore, we investigate how predictions of a trained network change when the landmark array is rotated. In \Cref{fig:gradcam}, plots A.1/B.1 and A.2/B.2 show predictions before and after rotation, respectively, with an overview of the rotation itself shown in the middle. For both A.2 and B.2, the network predicts a nest location that is shifted downwards by multiple meters. This shift seems to occur due to the network confusing the original center tree and rotated right tree (both marked $*$), with the latter closer to the nest. These results are not unlike those obtained from experiments performed by Collett and Cartwright~\cite{cartwright1983landmark}, where honey bees would maintain relative positions when landmark arrays were shifted.

\subsection{Control of a simulated quadrotor}
\label{ssec:control}

\begin{figure}[t!]
    \centering
    \includegraphics[width=0.95\linewidth]{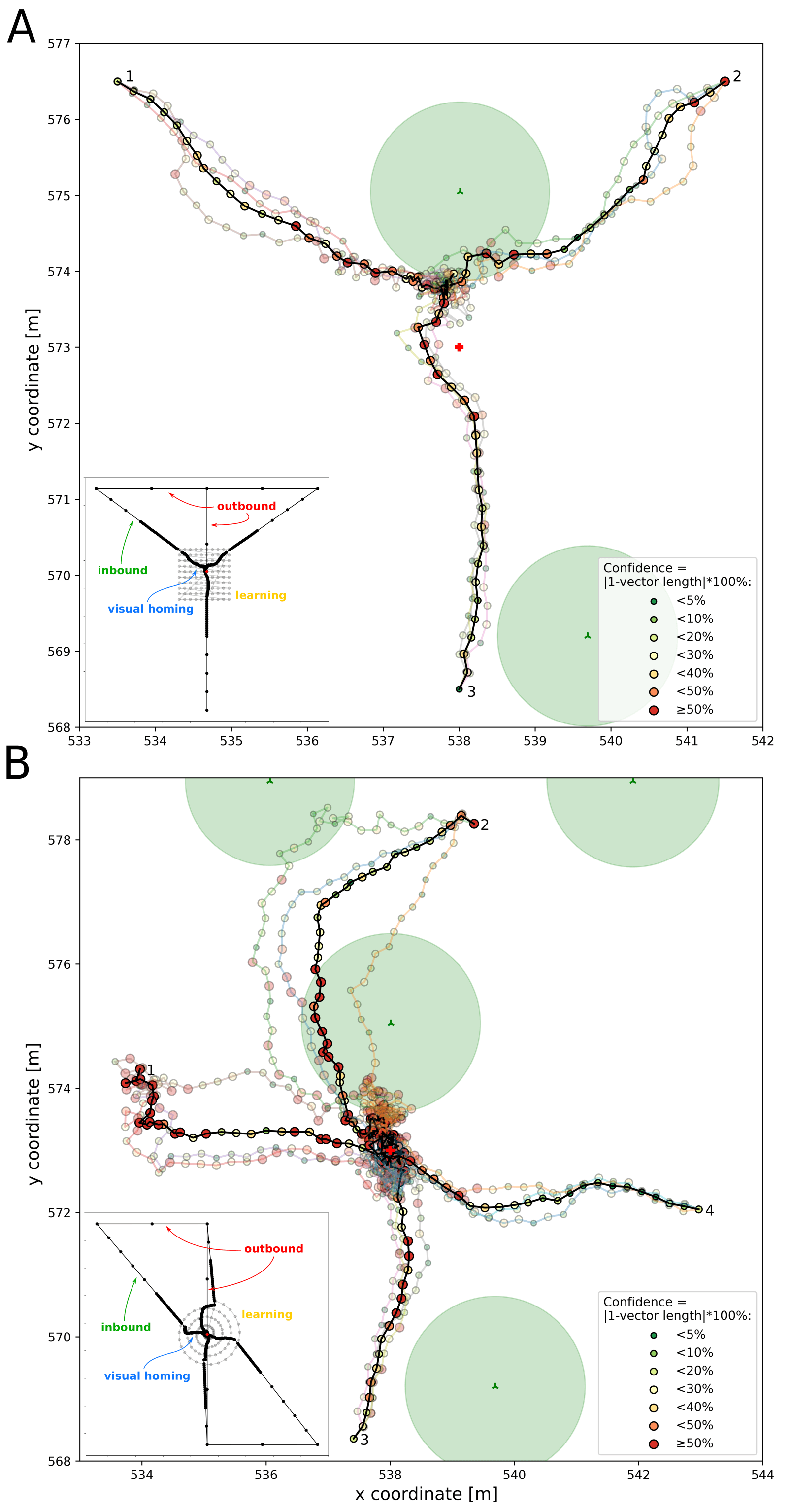}
    \caption{Homing performance when controlling a simulated quadrotor based on a network trained on (A) a $10\times10$~m grid and (B) an Archimedean spiral. Individual trajectories are shown transparent, with the mean trajectory per homing starting point in solid. Learning, outbound, inbound and visual homing phases are shown in the inset. Trees are shown in shaded green, and the nest is marked with a red cross at location $[538,573]^T$.}
    \label{fig:online}
\end{figure}

To further demonstrate generalization and usability, we use the output of the network to control a simulated quadrotor. First, we collect training data in either a grid or spiral learning flight and train the network as before. Second, the quadrotor embarks on an outbound flight. Third, we transition to an inbound phase, using path integration to fly in the nest direction. To simulate odometric drift, we stop the inbound flight at the edge of the learning trajectory. Fourth, we transition to visual homing, making use of the output of the network. The insets in \Cref{fig:online} show these steps for a single run per homing starting location, whereas the full figure shows the visual homing part of all runs (transparent trajectories, multiple starting locations).
Each run is made with an individually trained network that takes in an omnidirectional view at that location, sets the quadrotor's heading equal to the predicted home direction, and takes a fixed step of $0.25$~m in that direction, after which a new omnidirectional view is taken for the next step. If the quadrotor arrives within $0.25$~m of the nest location, the run is terminated and counted as a success.

As discussed in \Cref{ssec:generalization}, a too low sample density in the vicinity of the nest leads to an offset convergence point, and we see that in \Cref{fig:online}A as well. This means that none of the runs successfully reach the nest. On the other hand, when training on a spiral trajectory as in \Cref{fig:online}B, $10$ out of $12$ runs are successful, with on average $33.5$ steps taken. So, to ensure consistent returns, the learning trajectory in the vicinity of the nest should be of sufficiently high resolution.

\subsection{Real-world validation}

\begin{figure}[t!]
    \centering
    \includegraphics[width=\linewidth]{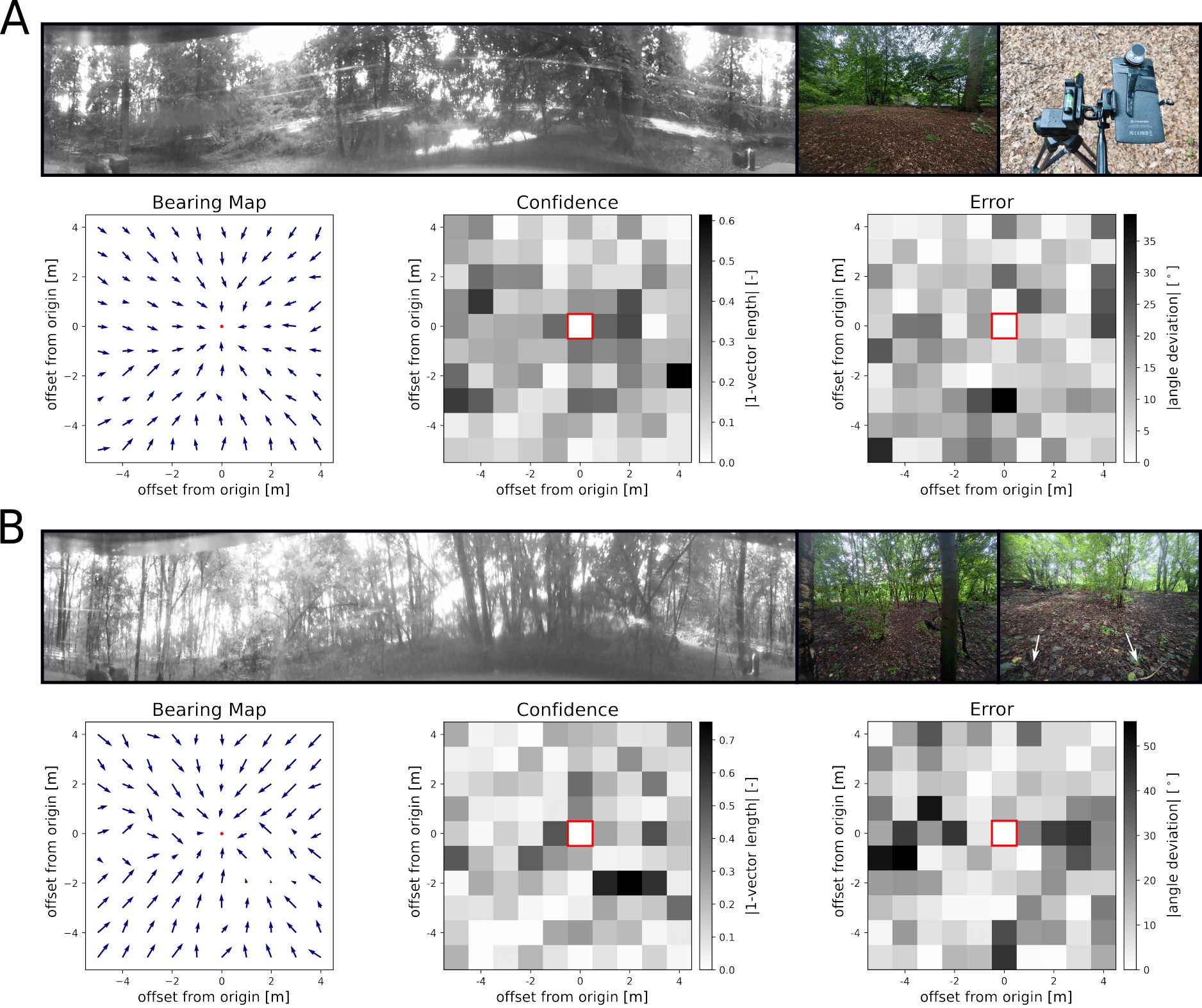}
    \caption{Training performance in two real forest environments, one (A) open and bright and another (B) covered and dark.
    The images show grayscale training samples and an overview of the experimental area, along with experimental equipment and sample location markers (white arrows). Nest locations are shown in red.}
    \label{fig:real}
\end{figure}

We show the robustness of our approach by validating it on data collected from $10\times10$~m grids in two real forest environments, shown in \Cref{fig:real}. Location A is relatively open and bright, while location B is denser and darker. Omnidirectional images are taken with a Oneplus One smartphone and a Kogeto Dot panoramic lens at a height of $\sim1.44$~m above the ground and a spacing of $1$~m.

As can be seen, the real-world images contain artifacts not present in their simulated counterparts; lens flare as well as no/inconsistent illumination make the real pictures unclear. Moreover, judging from the wave-like appearance of the images, the employed lens likely does not conform to the assumption of polar-to-Cartesian rectification. Despite all this, it seems that our approach transfers well to real-world data, with the bearing map and training error plot shown in \Cref{fig:real}A comparable to the ones in simulation (e.g., \Cref{fig:scene}A). In the case of location B, performance decreased due to the presence of more vegetation and shadows falling from the dense tree canopy on the forest floor. The respective average errors for locations A and B after training are $15.72^\circ$ and $30.52^\circ$, with the former again comparable to the $15.54^\circ$ obtained in simulation.

\section{DISCUSSION \& CONCLUSION}

We proposed a novel insect-inspired visual homing approach that relies exclusively on the currently experienced omnidirectional view. Our main hypothesis posits that bees learn directional cues to their nest during their learning flights, and use these cues to eliminate odometric drift when returning from a long foraging journey. We have subsequently demonstrated successful training of a compact convolutional neural network to predict the home direction in both simulated and real-world forest environments. Furthermore, we have investigated the generalization capability of such networks to locations not seen in training, as well as how they use environmental cues in their predictions. Successful control of a simulated quadrotor shows the promise of using our approach for visual homing on real robots.

This work can be extended in multiple ways, with first and most obvious the implementation of the network on a real robot with simplified 2D velocity control. Second, we observed that not shuffling the obtained images prior to training would lead to catastrophic forgetting. To allow true online during the learning trajectory (and not have to retain all images in memory), this will have to be mitigated~\cite{kirkpatrick2017overcoming}. Third, we envision to not only learn direction but also distance during the learning flights. This distance information can for example be used to adapt the flight velocity, allowing to speed up homing without losing homing precision. Lastly, to demonstrate the efficiency of our approach, we should compare it to existing map-building and relocalization techniques for robots like (visual) SLAM~\cite{kruzhkov2022meslam} and PoseNet~\cite{kendall2015posenet}.

\newpage
\bibliography{root}

\begin{thebibliography}{10}
\providecommand{\url}[1]{#1}
\csname url@rmstyle\endcsname
\providecommand{\newblock}{\relax}
\providecommand{\bibinfo}[2]{#2}
\providecommand\BIBentrySTDinterwordspacing{\spaceskip=0pt\relax}
\providecommand\BIBentryALTinterwordstretchfactor{4}
\providecommand\BIBentryALTinterwordspacing{\spaceskip=\fontdimen2\font plus
\BIBentryALTinterwordstretchfactor\fontdimen3\font minus \fontdimen4\font\relax}
\providecommand\BIBforeignlanguage[2]{{%
\expandafter\ifx\csname l@#1\endcsname\relax
\typeout{** WARNING: IEEEtran.bst: No hyphenation pattern has been}%
\typeout{** loaded for the language `#1'. Using the pattern for}%
\typeout{** the default language instead.}%
\else
\language=\csname l@#1\endcsname
\fi
#2}}

\bibitem{vonfrisch1993dance}
K.~{von Frisch}, \emph{The {{Dance Language}} and {{Orientation}} of {{Bees}}}.\hskip 1em plus 0.5em minus 0.4em\relax {Harvard University Press}, 1993.

\bibitem{haferlach2007evolving}
T.~Haferlach, J.~Wessnitzer, M.~Mangan, and B.~Webb, ``Evolving a {{Neural Model}} of {{Insect Path Integration}},'' \emph{Adaptive Behavior}, vol.~15, pp. 273--287, 2007.

\bibitem{srinivasan1996honeybee}
M.~V. Srinivasan, S.~W. Zhang, M.~Lehrer, and T.~S. Collett, ``Honeybee {{Navigation En Route}} to the {{Goal}}: {{Visual Flight Control}} and {{Odometry}},'' \emph{Journal of Experimental Biology}, vol. 199, pp. 237--244, 1996.

\bibitem{capaldi2000ontogeny}
E.~A. Capaldi, A.~D. Smith, J.~L. Osborne, S.~E. Fahrbach, S.~M. Farris, D.~R. Reynolds, A.~S. Edwards, A.~Martin, G.~E. Robinson, G.~M. Poppy, and J.~R. Riley, ``Ontogeny of orientation flight in the honeybee revealed by harmonic radar,'' \emph{Nature}, vol. 403, pp. 537--540, 2000.

\bibitem{zeil2012visual}
J.~Zeil, ``Visual homing: An insect perspective,'' \emph{Current Opinion in Neurobiology}, vol.~22, pp. 285--293, 2012.

\bibitem{cartwright1983landmark}
B.~A. Cartwright and T.~S. Collett, ``Landmark learning in bees,'' \emph{Journal of comparative physiology}, vol. 151, pp. 521--543, 1983.

\bibitem{lambrinos2000mobile}
D.~Lambrinos, R.~M{\"o}ller, T.~Labhart, R.~Pfeifer, and R.~Wehner, ``A mobile robot employing insect strategies for navigation,'' \emph{Robotics and Autonomous Systems}, vol.~30, pp. 39--64, 2000.

\bibitem{hafner2001learning}
V.~V. Hafner and R.~M{\"o}ller, ``Learning of {{Visual Navigation Strategies}},'' in \emph{Proceedings of the {{European Workshop}} on {{Learning Robots}}}, vol.~9, 2001, pp. 47--56.

\bibitem{baddeley2012model}
B.~Baddeley, P.~Graham, P.~Husbands, and A.~Philippides, ``A {{Model}} of {{Ant Route Navigation Driven}} by {{Scene Familiarity}},'' \emph{PLOS Computational Biology}, vol.~8, p. e1002336, 2012.

\bibitem{lemoel2020opponent}
F.~Le~M{\"o}el and A.~Wystrach, ``Opponent processes in visual memories: {{A}} model of attraction and repulsion in navigating insects' mushroom bodies,'' \emph{PLOS Computational Biology}, vol.~16, p. e1007631, 2020.

\bibitem{gattaux2023antcar}
G.~Gattaux, R.~Vimbert, A.~Wystrach, J.~R. Serres, and F.~Ruffier, ``Antcar: {{Simple Route Following Task}} with {{Ants-Inspired Vision}} and {{Neural Model}},'' 2023.

\bibitem{wehner1990insect}
R.~Wehner and S.~Wehner, ``Insect navigation: Use of maps or {{Ariadne}}'s thread?'' \emph{Ethology Ecology \& Evolution}, vol.~2, pp. 27--48, 1990.

\bibitem{webb2019internal}
B.~Webb, ``The internal maps of insects,'' \emph{Journal of Experimental Biology}, vol. 222, p. jeb188094, 2019.

\bibitem{chittka2022mind}
L.~Chittka, \emph{The {{Mind}} of a {{Bee}}}.\hskip 1em plus 0.5em minus 0.4em\relax {Princeton University Press}, 2022.

\bibitem{degen2015exploratory}
J.~Degen, A.~Kirbach, L.~Reiter, K.~Lehmann, P.~Norton, M.~Storms, M.~Koblofsky, S.~Winter, P.~B. Georgieva, H.~Nguyen, H.~Chamkhi, U.~Greggers, and R.~Menzel, ``Exploratory behaviour of honeybees during orientation flights,'' \emph{Animal Behaviour}, vol. 102, pp. 45--57, 2015.

\bibitem{vandalen2018visual}
G.~J.~J. {van Dalen}, K.~N. McGuire, and G.~C. H.~E. {de Croon}, ``Visual {{Homing}} for {{Micro Aerial Vehicles Using Scene Familiarity}},'' \emph{Unmanned Systems}, vol.~06, pp. 119--130, 2018.

\bibitem{song2021flightmare}
Y.~Song, S.~Naji, E.~Kaufmann, A.~Loquercio, and D.~Scaramuzza, ``Flightmare: {{A Flexible Quadrotor Simulator}},'' in \emph{Proceedings of the 2020 {{Conference}} on {{Robot Learning}}}.\hskip 1em plus 0.5em minus 0.4em\relax {PMLR}, 2021, pp. 1147--1157.

\bibitem{berenguel-baeta2020omniscv}
B.~{Berenguel-Baeta}, J.~{Bermudez-Cameo}, and J.~J. Guerrero, ``{{OmniSCV}}: {{An Omnidirectional Synthetic Image Generator}} for {{Computer Vision}},'' \emph{Sensors}, vol.~20, p. 2066, 2020.

\bibitem{gould1978bees}
J.~L. Gould, J.~L. Kirschvink, and K.~S. Deffeyes, ``Bees {{Have Magnetic Remanence}},'' \emph{Science}, vol. 201, pp. 1026--1028, 1978.

\bibitem{rossel1986polarization}
S.~Rossel and R.~Wehner, ``Polarization vision in bees,'' \emph{Nature}, vol. 323, pp. 128--131, 1986.

\bibitem{selvaraju2017gradcam}
R.~R. Selvaraju, M.~Cogswell, A.~Das, R.~Vedantam, D.~Parikh, and D.~Batra, ``Grad-{{CAM}}: {{Visual Explanations From Deep Networks}} via {{Gradient-Based Localization}},'' in \emph{Proceedings of the {{IEEE International Conference}} on {{Computer Vision}}}, 2017, pp. 618--626.

\bibitem{kirkpatrick2017overcoming}
J.~Kirkpatrick, R.~Pascanu, N.~Rabinowitz, J.~Veness, G.~Desjardins, A.~A. Rusu, K.~Milan, J.~Quan, T.~Ramalho, A.~{Grabska-Barwinska}, D.~Hassabis, C.~Clopath, D.~Kumaran, and R.~Hadsell, ``Overcoming catastrophic forgetting in neural networks,'' \emph{Proceedings of the National Academy of Sciences}, vol. 114, pp. 3521--3526, 2017.

\bibitem{kruzhkov2022meslam}
E.~Kruzhkov, A.~Savinykh, P.~Karpyshev, M.~Kurenkov, E.~Yudin, A.~Potapov, and D.~Tsetserukou, ``{{MeSLAM}}: {{Memory Efficient SLAM}} based on {{Neural Fields}},'' in \emph{2022 {{IEEE International Conference}} on {{Systems}}, {{Man}}, and {{Cybernetics}} ({{SMC}})}, 2022, pp. 430--435.

\bibitem{kendall2015posenet}
A.~Kendall, M.~Grimes, and R.~Cipolla, ``{{PoseNet}}: {{A Convolutional Network}} for {{Real-Time}} 6-{{DOF Camera Relocalization}},'' in \emph{Proceedings of the {{IEEE International Conference}} on {{Computer Vision}}}, 2015, pp. 2938--2946.

\end{thebibliography}
\bibliographystyle{IEEEtran}

\end{document}